\pdfoutput=1
\documentclass[11pt]{article}

\usepackage[]{EMNLP2022}

\usepackage{times}
\usepackage{latexsym}
\usepackage{graphicx}
\usepackage{makecell}
\usepackage{multirow}
\usepackage{diagbox}
\usepackage{droidsansmono}
\usepackage{enumitem}

\usepackage[T1]{fontenc}
\usepackage[utf8]{inputenc}

\usepackage{microtype}

\usepackage{inconsolata}

\definecolor{custom_pink}{RGB}{227, 129, 129}
\definecolor{custom_cyan}{RGB}{118, 178, 223}

\title{CodeExp: Explanatory Code Document Generation}

\author{Haotian Cui$^1$\Thanks{ Work was done during an internship at Microsoft.}, Chenglong Wang$^2$, Junjie Huang$^3$, Jeevana Priya Inala$^2$\\
    \bf{Todd Mytkowicz$^2$, Bo Wang$^1$, Jianfeng Gao$^2$, Nan Duan$^2$\Thanks{ Corresponding author.}} \\
    $^1$University of Toronto, $^2$Microsoft Research, $^3$Beihang University \\
    {\{ht.cui, bowang.wang\}@mail.utoronto.ca,}\\
    { huangjunjie@buaa.edu.cn} \\
    { \{chenwang, jinala, toddm, jfgao, nanduan\}@microsoft.com,}\\
  }

\begin{document}
\maketitle
\begin{abstract}
Developing models that can automatically generate detailed code explanation can greatly benefit software maintenance and programming education. However, existing code-to-text generation models often produce only high-level summaries of code that do not capture implementation-level choices essential for these scenarios. To fill in this gap, we propose the \emph{code explanation generation} task. We first conducted a human study to identify the criteria for high-quality explanatory docstring for code. Based on that, we collected and refined a large-scale code docstring corpus and formulated automatic evaluation metrics that best match human assessments. Finally, we present a multi-stage fine-tuning strategy and baseline models for the task. Our experiments show that (1) our refined training dataset lets models achieve better performance in the explanation generation tasks compared to larger unrefined data (15$\times$ larger), and (2) fine-tuned models can generate {well-structured long docstrings} comparable to human-written ones. We envision our training dataset, human-evaluation protocol, recommended metrics, and fine-tuning strategy can boost future code explanation research. The code and annotated data are available at \url{https://github.com/subercui/CodeExp}.
% Automatic code explanation is valuable for program learning and code maintenance. However, existing code summarization methods do not explain code semantics in detail. In this work, we examine the uniqueness and challenges of explanatory document generation. We propose evaluation criteria for the task, collect large-scale high-quality code docstring corpus boosted with human annotations, and select automatic evaluation metrics best matching human assessment. We highlight the importance of data quality by showing that fine-tuning on high-quality examples exceeds the performance using 15 times larger-scale data of mixed qualities. The fine-tuned models can generate decent \textbf{well-formatted long docstrings} comparable to human-written ones. We expect the proposed infrastructures, including the annotated dataset, human-evaluation protocol, recommended metrics, and fine-tuning strategy, to boost future research for code explanation.

\end{abstract}

\section{Introduction}
% a.	Explaining code is important, for learners, maintainers. Cases: Denigma and GitHub Copilot Labs
% b.	limited studies for code-explanation so far. Code-summary task is different. A figure of example showing the difference between summary and explanation.
% Summarize the code-summary and explanation. E.g. , lack of evaluation metrics.
% c.	Contributions: (1) propose task. (2) Benchmark: large auto-filtered dataset for fine-tuning; labeled dataset for evaluation. (3) Benchmark: Human-eval set-up. Compare and propose the proper combination of auto-metrics accordingly. (4) Base-line models. New fine-tuning set-up (stage 1 and stage 2 fine-tuning) shows promising results.

\begin{figure}[t]
    \centering
    \includegraphics[width=0.48\textwidth]{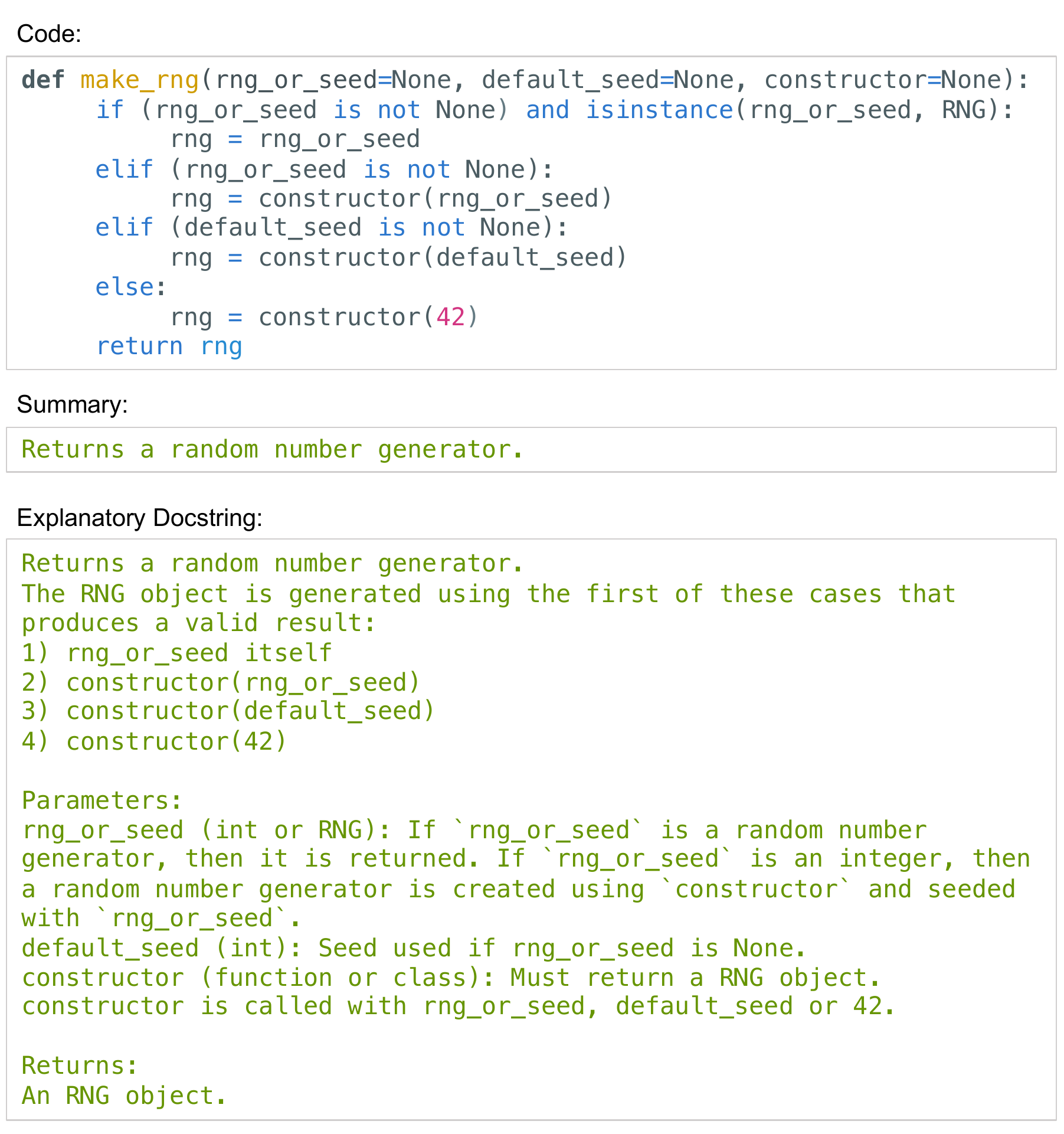}
    \caption{Example code paired with summary and explanatory docstring. Difference between two styles: Summaries outline the highest-level intent of the code. Docstrings are more \textbf{informative and detailed}, explaining the semantics of specific code pieces.}
    \label{fig:example}
\end{figure}

Code documentation improves program comprehension \citep{garousi2015usage} and reduces software maintenance cost~\citep{chen2009empirical}. %However, manually documenting software is a tedious task.
%However, despite its importance, many software are under-documented because creating code documentation can be tedious \chenglong{add citation}.
Recently, many automated code summary tools have been developed to reduce the effort of document creation: for example, Denigma\footnote{\url{https://denigma.app/}} is an IDE extension for generating inline function summaries; GitHub Copilot Labs\footnote{\url{https://github.com/github/feedback/discussions/8308}} is a code summary model built on top of Codex~\citep{Chen2021} for generating explanations for AI generated code. % and is expected to be added into the copilot service in near future.
However, these existing code summary tools focus on the generation of short high-level descriptions of source code semantics~\citep{haiduc2010supporting, roy2021reassessing, zhu2019automatic}, and code summary alone is insufficient to meet the software understanding and maintenance need: %practical code documentations also require detailed explanatory documents that explain implementation choices and logic behind the code \cite{tenny1988program, das2007understanding}. 
A recent survey shows 85\% developers expect tools to generate method-level documentations explaining the functionalities, usage and design rationales of code \cite{hu2022practitioners}. % ~\chenglong{add some citation to justify this?}. 
%  few studies have explored generating long-form detailed code explanations to the best of our knowledge. Recent efforts mainly focused on automatic code summarization, i.e.
For example, as shown in Figure~\ref{fig:example}, the code summary captures only the high-level code functionality, while the explanatory docstring explains arguments, return values, and its computation process in a detailed and informative way. %which is essential for maintainers or code learners efficiently understand the internal logic of the function.
However, due to the lack of training and evaluation resources, few code explanation models have been developed.

% However, existing code summarization studies focus on the generation of very high-level description of code logic, usually within one sentence.
% We argue that there is a large gap between the current generated summaries and the aforementioned explanatory documentation. An example is shown in Figure~\ref{fig:example}. 
% In contrast to short summaries, the explanatory document needs to be \textbf{informative} and \textbf{detailed}, to help the maintainers or code learners efficiently understand the internal logic of code pieces.

% some where mention, we choose to work with python
In this work, we introduce the {\bf code explanation generation} task. We provide (1) the training corpus, (2) fine-tuning strategy, and (3) human-evaluation protocol and recommended automatic evaluation metrics to support developing code explanation models. Our contributions include:
% 1. data 2. models 3. metrics
{\setlist{leftmargin=5pt}
\begin{itemize}\itemsep0pt
    \item We provide a python code-docstring corpus \textit{CodeExp}, which contains (1) a large partition of 2.3 million raw code-docstring pairs, (2) a medium partition of 158 thousand pairs refined from the raw corpus using a learned filter, and (3) a partition of 13 thousand pairs with rigorous human annotations. Our data collection process leverages an annotation model learned from human annotations to automatically filter high quality code-docstring pairs from raw GitHub datasets.
    \item We propose a two-stage strategy for fine-tuning large language models using collected data --- first with the raw data and then with medium-size refined data. Our experiments show that the best fine-tuned model achieves human-comparable performance and highlights the importance of high-quality data for the code explanation task.
    \item We evaluated our models on seven automatic evaluation metrics and examined their consistency with respect to the human evaluation of 180 test examples. Our study shows that BLEU \cite{papineni2002bleu} and METEOR \cite{banerjee2005meteor} best reflect code generation quality, and we recommend using them in future research.
\end{itemize}}

\section{Code Explanation Generation}
\label{sec:task}
\paragraph{Application scenarios.} We focus on the generation of code explanations that describe both low-level and high-level code semantics. The automatic code explanation tool can benefit developers in many scenarios. For example, the tool can reduce software engineers developing effort by automatically generating function comments during development; it can help learners and codebase maintainers better understand undocumented code; it can also explain code generated by code generation models like Codex for the developer to better understand and verify its correctness. Note that in these scenarios, because the developer needs to understand both design rationale and implementation details of the code, an explanation covering detailed code semantics would be more appropriate than a short high-level description. For example, if a developer aims to create a test for the function $\mathsf{make\_rng}$ in Figure~\ref{fig:example} when maintaining a codebase, it is crucial to understand how the variable $\mathsf{rng\_or\_seed}$ is defined and used. 

\paragraph{Task definition.} Based on these observations, we define the code explanation task as the text generation given code snippets (functions), where the generated texts describe the code semantics. Concretely, a high quality code explanation should meet the following criteria: the description should be informative, covering important code behaviors, coherent with the semantics of source code, fluent, and grammatically correct. We follow these criteria to set up our annotation and evaluation protocols in Section~\ref{sec:human-ann-data} and \ref{sec:human-eval}, respectively.

%Note that code summarization can be considered as a special instance where only the high-level semantics is included.

\paragraph{Challenges.}  The first key challenge for developing code explanation models is the shortage of high-quality paired training and evaluation data. Despite the existence of large-scale public code corpora, directly using functions and their comments for modelling is not ideal because these comments can be oversimplified or misleading: \citet{clement2020pymt5} found 44\% of python function documents are very short in one-line style, and \citet{wen2019large} showed code changes rarely (<20\%) trigger comments updates, potentially making the inconsistency between code and comments a severe issue.

Second, developing code explanation models is challenging. Besides the need for understanding code logic to generate a {global} summary, the explanation model also needs to generate detailed comments based on fine-grained local code structures (e.g., examine the control and data flow in order to explain how a variable is used). %This brings up new modeling opportunities. %\citet{xxx} also showed that the function identifiers often contain sufficient hints to make a good summary.  
%For example, to understand the effect of $\mathsf{rng\_or\_seed}$ in Figure~\ref{fig:example}, one needs to look at the specific if/else conditions and connect these findings across several code lines. Overall, generating code explanations demands a combination of both \textbf{global and local} understanding of codes. 
%This capability is one of the key challenges, and it has been hardly verified in code summarization studies.

Third, while prior work \cite{gros2020code, roy2021reassessing} empirically studied criteria of high quality code summaries, translating these criteria/guidelines into actionable automatic evaluation metrics for code explanation remains a challenge.

%Given the scarcity of high-quality data, function-level comments are the best existing large data source to our best knowledge. Without the loss of generality, we focus on python functions and document strings (docstrings).

 %To relieve the shortage of paired data and provide training and evaluation infrastructure for the code explanation task, we set up human-evaluation protocol and collect large amount of high-quality code docstring pairs (Section~\ref{sec:data}); we fine-tuned baseline models (Section~\ref{sec:exp-setup}) and discussed suitable evaluation metrics(Section~\ref{sec:human-eval}). Eventually we discussed a proper fine-tuning and evaluation strategy for the task of code explanation (Section~\ref{sec:result}).

We next present how we collect high quality datasets, define evaluation metrics and fine-tune language models to address the above challenges.

% \section{CodeExp}
\section{CodeExp Data Collection}
\label{sec:data}
% a.	Human-labeled 13k data: labeling set-up; data source; annotators and workflow description; result data statistics; What can be done with the data and annotations.
% b.	Training ml annotators to mimic human labelers.
% c.	2.2 million data pairs from GitHub: selection criteria; Processing workflow; Usage-stage 1 fine-tuning (May be used for pretraining, not shown in this paper). 
% d.	160 k filtered data pairs by ml annotators. Filtering criteria; conceptually high-quality data for stage 2 finetuning.

This section describes our data selection process. The code explanation corpus (\textit{CodeExp}) consist of three sets of code-docstring pairs: (1) CodeExp(annotated) with 13K human annotated pairs. (2) CodeExp(refined) with 158K pairs filtered by a trained model. (3) CodeExp(raw) with 2.3M unlabelled pairs.

\subsection{Examine docstring quality with human annotations - CodeExp(annotated)}
\label{sec:human-ann-data}
In order to understand how developers evaluate the quality of explanatory docstrings, we first conduct a user study to let developers annotate quality of code-docstring pairs using the code-doc-corpus collected by \citet{barone2017parallel}. The corpus contains 109,108 parallel python functions and docstrings. We automatically filtered the dataset to only include qualified examples. The filtered examples consist of the data pairs where (1) the number of code lines is within 6 to 30, (2) the number of lines in the docstring is larger than 3, and (3) the Cyclomatic Complexity\footnote{\url{https://radon.readthedocs.io/en/latest/intro.html\#cyclomatic-complexity}} of the code is larger than 3. In total, this yields 13,186 valid examples. We hire external annotators to label this subset.

\paragraph{Annotation} Our annotation protocol is developed through several pilots and further updated with hard examples as the annotation progresses. 
Annotators are asked to make three judgements for each pair of code and docstring: (1) \textbf{General adequacy:}  The docstring should describe the main logic of the code, i.e. containing at least one sentence describing how the code handles the input or what it computes given the input. (2) \textbf{Coverage:} If the code contains outer-level if/else or try/except blocks, check whether the docstring describes the semantic of each block. (3) \textbf{Coherence:} Check whether the documentation (if any) of parameter types and returns match the code semantics.

\begin{figure}[h]
    \centering
    \includegraphics[width=0.45\textwidth]{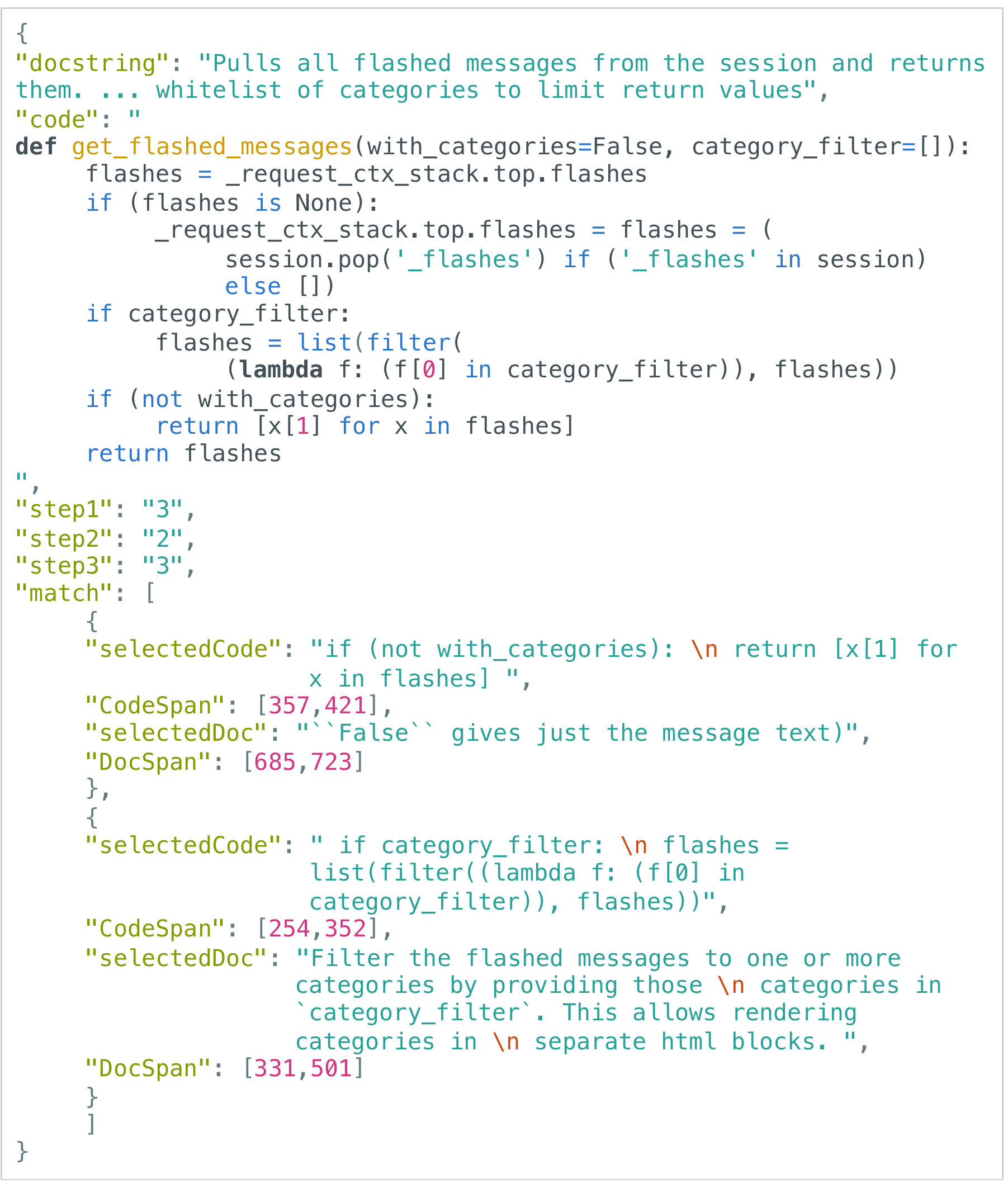}
    \caption{An annotated example of high-score code docstring pair. Integer scores of three steps are provided. The "CodeSpan" field extracts the specific code lines described by the texts of the "DocSpan" field.}
    \label{fig:data_example}
\end{figure}

For each pair of code and docstring, the human annotator is asked to give a score from 0 (worst) to 3 (best) for each step if it is applicable (only step~1 is always applicable); otherwise, the annotator leaves a blank score. For the coverage evaluation, the annotator will also mark the specific code spans and text spans that are associated with code blocks (Figure~\ref{fig:data_example}). Due to the concern of feasibility, we do not require the step~2 annotation to cover all aspects of details, but only the branching blocks. %\junjie{can we just leave the in-applicable ones to score 0? }

We refer to the human-annotated data as the \textbf{CodeExp(annotated)} in later sections. We show the statistics of annotations in Appendix~\ref{sec:ann-stats}. In the annotated result, we found the human-written docstrings mostly perform well in explaining the general logic and provide accurate type defines, but are less optimal considering the coverage requirement. In other words, examples have high scores for step~1 and step~3, but the scores for step~2 interestingly diverge. For the 11,900 code with branching blocks, 6,300 examples do not describe any block (step~2 score equals zero). The rest 5,400 examples (33\%) describe at least one code block.

%show example of the annotated code-doc pair.

\subsection{CodeExp(raw) with Open-source Pairs}
As it is economically infeasible to manually annotate a dataset large enough with code-docstring pairs for training a machine learning model, we instead create a suboptimal dataset with unlabelled pairs. 
Following \cite{husain2019codesearchnet}, we pair the code function with its corresponding documentation to form a code-docstring pair. 
We leverage open-source python repositories in GitHub to collect CodeExp(raw).
To ensure the code quality, we only keep repositories with more than 60 stars. This step yields around 55,000 repositories by December 2021.
We downloaded all Python files with \textit{'.py'} extensions and parsed the source code into abstract syntax trees (AST) with Tree-Sitter\footnote{\url{https://tree-sitter.github.io/tree-sitter/}}. Then we include the functions provided with docstrings. 
Finally, we collected a corpus of 2,285,387 pairs of Python function code and docstrings. 
To the best of our knowledge, CodeExp(raw) is one of the largest datasets with parallel programming language (PL) and natural language (NL) till now. % which is 5x larger than previous largest one CodeSearchNet \cite{husain2019codesearchnet}.
In comparison, CodeSearchNet \citep{husain2019codesearchnet} contains 457,461 samples of python code.

\subsection{Data selection with a learned filter - CodeExp(refined)}
\label{sec:data-select}
The collected code and docstrings from Github are of mixed qualities and potentially introduce noise if used for training. Hence, we aim to refine a higher-quality subset from it for modeling. Our key insight here is to train a machine learning filter to mimic the human annotators based on the annotated dataset, and apply the learned filter to refine the raw data. The filter is fine-tuned from a pretrained BERT base model (uncased) on the collected human annotations. It takes as input the code and docstring pairs and predicts the step 1 and 2 scores \footnote{Because most annotated examples have high step 3 scores, and low-score examples are inconsistent between annotators, we excluded step 3 score when training the ML filter.}. The target scores are normalized to $[0,1]$. We used 11208 examples, 85\% of CodeExp(annotated), for training and the rest for validation. The model achieved mean square errors (mse) of 0.027 and 0.018 for step 1 and 2, respectively, on the validation set.

We apply the same workflow as in Section~\ref{sec:human-ann-data} to filter the raw 2.3M corpus of CodeExp(raw): we used the same complexity and length threshold to select candidate examples, and then applied the ML-filter. Finally, we selected the qualified data pairs with predicted step 1 and 2 scores greater than 1.0 (after scaling back). In this way, we collected 158,024 refined examples, named as the \textbf{CodeExp(refined)} partition.

\begin{table}[h]
\centering
    \resizebox{0.48\textwidth}{!}{
        \begin{tabular}{llll}
        \hline
        \textbf{Partition} & \textbf{\#Examples} & \textbf{Quality} & \textbf{Annotated By}\\
        \hline
        CodeExp(raw) & 2,285,387 & Mixed & -  \\
        CodeExp(refined) & 158,024 & High & Machine\\
        CodeExp(annotated) & 13,186 & Mixed & Human \\
        \hline
        \end{tabular}
    }
    \caption{\label{table:data} Data statistics. The raw and refined partitions are collected from \href{https://github.com}{GitHub}. The annotated partition is selected from code-doc-corpus \citep{barone2017parallel}.
    }
\vspace{-1mm}
\end{table}

Table~\ref{table:data} shows the statistics for the three partitions. Note that the annotated subset's quality is ``mixed'' because it contains both low and high scored examples annotated by human (we only use higher scored ones for testing in Section~\ref{sec:finetune}). Because the refined partition is refined by the learned filter, it better matches the definition of code explanations (Section~\ref{sec:task}). An example with human annotations is shown in Figure~\ref{fig:data_example}. 

\section{Experiment Settings}
\label{sec:exp-setup}
% a.	Three choices for fine-tuning (1) on 2.2 million (2) on 160k (1,2) on both
% b.	Baseline models
% The docstring generation is formed as a sequence-to-sequence task. We use code-only functions (including function definition headers and function bodies, excluding inline comments) as the source texts, and developer-written docstrings as the target/reference sequences. 
In this paper, we formulate the code explanation generation as a sequence-to-sequence problem, where the source sequence is a function code (including both function signature and body), and the target sequence is an explanatory documentation string.
We select four strong pretrained language models for programming language and fine-tune the models on our proposed CodeExp dataset.
We next introduce the baseline models, experiment settings, and evaluation metrics. 

\subsection{Baseline models}
\label{sec:model}
We evaluate four popular pretrained language models in this paper:  (1) \textit{GPT2-base} \citep{radford2019language}, a transformer decoder model pretrained with only NL; (2) \textit{GPT-Neo} \citep{gpt-neo},  a series of GPT-style decoders varying in parameter size and pretrained with the large PILE corpus \cite{gao2020pile} containing both NL and PL;
(3) \textit{CodeT5} \cite{wang2021codet5}, a transformer encoder-decoder model pretrained with NL and PL. (4) \textit{Codex} \cite{chen2021evaluating} the state-of-the-art model trained on Github code of multiple programming languages including python.
We fine-tune these models with our collected data to test the performance except for Codex, with which we only report its zero-shot performance due to the inaccessibility of model weights. 
% We select four baseline models:  \textit{GPT2-base} \citep{radford2019language}, \textit{GPT-Neo} \citep{gpt-neo}, \textit{CodeT5} \cite{wang2021codet5} and Codex \cite{chen2021evaluating}. Notably, these models vary in parameter size (Table~\ref{table:model statistics}), architecture, and pretrained data: \textit{GPT-2-base}, a transformer decoder model, was pretrained with NL only; \textit{GPT-Neo} series are GPT-style decoders pretrained with the large PILE corpus \cite{gao2020pile}, which contains both NL and PL; \textit{CodeT5} consists of both encoder and decoder modules and was trained mainly with PL. The only NL data used for training \textit{CodeT5} were code comments, which were much more domain-specific than the NL texts used for \textit{GPT-2} and \textit{GPT-neo}. Codex\cite{chen2021evaluating}

\begin{table}
\centering
    \resizebox{0.48\textwidth}{!}{
        \begin{tabular}{lll}
        \hline
        \textbf{Models} & \textbf{\#Params} & \textbf{Pretrained w/} \\
        \hline
        GPT-2-base \citep{radford2019language} &  117M & NL\\
        GPT-Neo-13 \citep{gpt-neo} & 1.3B & NL, PL \\
        GPT-Neo-27 \citep{gpt-neo} & 2.7B & NL, PL \\
        CodeT5 \citep{wang2021codet5} & 220M & PL$^*$\\
        \hline
        \end{tabular}
    }
\caption{\label{table:model statistics} Backbone models for fine-tuning. ($^*$) CodeT5 was mainly trained with PL. The NL training data contained only the code comments.}
\end{table}

\subsection{Fine-tune}
\label{sec:finetune}
We fine-tune the baseline models with the collected data of different sizes and qualities (Table~\ref{table:model statistics}). This yields three fine-tuning strategies: \textbf{S1.} Fine-tune on all collected examples, i.e. CodeExp(raw); \textbf{S2.} Fine-tune on the refined subset, CodeExp(refined); \textbf{S3.} Fine-tune on the raw and refined partitions consecutively, in a "curriculum learning" manner. For each strategy, we leave out 1\% of the raw partition and/or 5\% of the refined partition for validation. 

For evaluation, we select examples with high scores from the CodeExp(annotated) and remove duplicates if they also appear in the raw or refined partition (Appendix~\ref{sec:test-set}). This generates a high-quality test set of 2,677 examples.

The pretrained models are fine-tuned using cross-entropy loss on 16 Nvidia V100 32GB GPUs. We select the checkpoint with the best perplexity score on the corresponding validation set for further evaluation. For strategy S3, the best checkpoint after fine-tuning on CodeExp(raw) is used as the starting point for the next phase on CodeExp(refined). The hyperparameters, including the max tokens, learning rate, batch size, epoch numbers, etc., are listed in Appendix Table~\ref{tab:hyper-param}. At inference time, we sample the top generated text using the default inference settings (Appendix~\ref{sec:inference}).

\subsection{Automatic metrics}
\label{sec:auto-metric}
We adopt four existing metrics  widely used in related tasks and propose two new metrics dubbed \textit{CER} and \textit{CodeBERTScore} to evaluate our models. 
The efficacy of each metric is verified in the next section against human evaluation. 
We include both statistical and recent model-based metrics. 

\paragraph{Statistical Metrics} BLEU \cite{papineni2002bleu}, ROUGE \cite{lin2004automatic}, METEOR \cite{banerjee2005meteor} are three commonly used metrics in code summarization and machine translation. These methods compute the matching of n-grams between candidate and target texts in various manners. We use the sentence BLEU (with smoothing method 4) and METEOR interfaces provided by NLTK\footnote{\url{https://www.nltk.org/}}. For ROUGE, we use the F1 score of ROUGE-1 and ROUGE-L.

We propose a new metric dubbed Common Entity Recall (\textit{CER}). It first computes the number of common 1-grams shown in code, generated, and reference docstrings. Then it is divided by the number of common 1-grams of the code and reference docstring. The intuition is that we found the common 1-grams of the code and reference docstrings often contain important variable names, function identifiers, or important keywords (e.g., if, int).

\paragraph{Model-based Metrics} BERTScore \cite{zhang2019bertscore} is an automatic metric that employs a BERT model to measure the similarity between generation and reference. The semantic similarity is  computed as cosine similarities between the average token embeddings of generated and target texts.

As BERTScore is only for natural language, we propose CodeBERTScore as an adapted version of BERTScore for evaluating code-related tasks. This metric is built by replacing the language model in BERTScore with CodeBERT \cite{feng2020codebert}, a state-of-the-art language model pretrained with code and natural language. We use the average token embeddings of the 9th layer in CodeBERT to compute similarity.

\section{Metric selection based on human evaluation} % or call this Human Evaluation
\label{sec:human-eval}

We conducted human evaluations on generated docstrings and compared the aforementioned automatic metrics against the human-eval results. The protocol of our human evaluation is designed to cover four important aspects: \textbf{A1}. General adequacy, \textbf{A2.} Coverage, \textbf{A3.} Coherence, \textbf{A4.} Fluency. % (5) Docstring style. 
Annotators rate for each aspect a score within 0-4 on the 5-point Likert scale \cite{likert1932technique}. The detailed setup is listed in Appendix~\ref{sec:human-eval-setting}. Aspects 1,3,4 have been used in both machine translation \cite{reiter2018structured} and code summarization studies \cite{song2019survey, roy2021reassessing}. The coverage aspect emphasizes the preference for informative explanations of code pieces. %The docstring style score depicts how well the generated docstring follows the docstring conventions of PEP257 \footnote{\url{https://peps.python.org/pep-0257/}}.
Notably, these scores provide reference-free assessments. Both the original reference docstring (identity is hidden) and generated ones are provided to annotators.

We calculate the adapted Kendal's $\tau$ \cite{graham2015accurate} to measure the agreement between automatic metrics and human evaluation. For an arbitrary pair of two examples, it considers whether two metrics both prefer one example to the other. The $\tau$ value is the ratio difference of concordant (Con) and discordant (Dis) pairs,
\begin{equation}
    \tau = \frac{|\texttt{\#Con} - \texttt{\#Dis}|}{|\texttt{\#Con} + \texttt{\#Dis} + \texttt{\#Tie}|},
\end{equation}
where $\texttt{\#}$ denotes the number of pairs. The concordant, discordant and tie pairs are calculated as in Table~\ref{tab:concordant}, where $s_1$ and $s_2$ are the scores for the two docstrings within a pair.

\begin{table}[h]
\centering
    \resizebox{0.45\textwidth}{!}{
        \begin{tabular}{cc|ccc}
                                    &                 & \multicolumn{3}{c}{Metric}  \\
                                    &                 & $s_1 < s_2$ & $s_1 = s_2$ & $s_1 > s_2$ \\ 
        \hline
        \multirow{3}{*}{Human}      & $s_1 < s_2$ & concordant & tie   & discordant                  \\
                                    & $s_1 = s_2$ & -  & -  & -                 \\
                                    & $s_1 > s_2$ & discordant & tie   & concordant                 
        \end{tabular}
    }
    \caption{Calculate concordant and discordant pairs.}
    \label{tab:concordant}
\end{table}

% Several previous code summarization works have employed Kendal's $\tau$ to study the entailment between automatic metrics and human evaluations \cite{graham2015accurate, roy2021reassessing}, where automatic metrics showed $\tau$ values within 0.25-0.5.

% two research questions.

% \textbf{RQ1.} To what extend do auto-matic metrics agree with the human evaluation?
% % Do the metrics reflect human assessments?

% \textbf{RQ2.} What metrics can be adopted as proxies of human evaluation for docstring generation?
% % apply auto-metrics from other seq2seq

\begin{table*}[h]
\centering
    \resizebox{0.86\textwidth}{!}{
        \begin{tabular}{lccccccc}
        \hline
        \textbf{Model} & \textbf{ROUGE-1 f} & \textbf{ROUGE-L f} & \textbf{BLEU}  & \textbf{CER} & \textbf{METEOR} & \textbf{BERTScore} & \makecell[c]{\textbf{CodeBERT}\\\textbf{Score}} \\ %& \makecell[c]{\textbf{Combined}\\\textbf{Score}} \\
        \hline
        GPT-2-base           &           &           &       &               &        &           &     \\
        \hspace{0.3cm}- CodeExp(raw)     & 0.2557    & 0.2443    & 4.42  & 0.4580        & 18.55  & 83.87    & 77.46     \\
        \hspace{0.3cm}- CodeExp(refined)     & 0.2462    & 0.2357    & 4.26  & 0.4890        & 19.00  & 83.55    & 77.48     \\
        \hspace{0.3cm}- CodeExp(r+r)   & 0.2623    & 0.2520    & 5.19  & 0.4978       & 20.30  & 83.92    & 77.91     \\
        \hline
        GPT-Neo13           &           &           &       &               &        &           &               \\
        \hspace{0.3cm}- w/o fine-tune & 0.0694    & 0.0646    & 0.40  & 0.1294        & 5.71   & 78.19    & 67.62         \\
        \hspace{0.3cm}- CodeExp(raw)   & 0.2894    & 0.2769    & 7.51 & 0.4805        & 21.85  & 84.46    & 78.31   \\
        \hspace{0.3cm}- CodeExp(refined)   & 0.2954    & 0.2815    & 7.36 & 0.5570        & 23.58  & 84.24    & 78.46   \\
        \hspace{0.3cm}- CodeExp(r+r) & \textbf{0.3265}    & \textbf{0.3128}    & \textbf{10.45} & 0.5524        & \textbf{26.31}  & \textbf{84.78}    & \textbf{79.26}   \\
        \hline
        GPT-Neo27           &           &           &       &               &        &           &               \\
        \hspace{0.3cm}- w/o fine-tune & 0.1480    & 0.1380    & 0.82  & 0.2285        & 10.06   & 78.40    & 71.99         \\
        \hspace{0.3cm}- CodeExp(raw)   & 0.2953    & 0.2818    & 7.72 & 0.4895        & 22.42  & 84.31    & 78.34   \\
        \hspace{0.3cm}- CodeExp(refined)   & 0.2955    & 0.2816    & 8.21 & 0.5285        & 23.66  & 84.33   & 78.41   \\
        \hspace{0.3cm}- CodeExp(r+r) & \textbf{0.3298}    & \textbf{0.3154}    & \textbf{\underline{10.72}} & \textbf{0.5560}        & \textbf{26.87}  & \textbf{84.86}    & \textbf{79.28}     \\
        \hline
        CodeT5           &           &           &       &               &        &           &               \\
        \hspace{0.3cm}- multi-sum    & 0.1507    & 0.1392    & 0.21  & 0.1240        & 5.58   & 82.35    & 74.29   \\
        \hspace{0.3cm}- CodeExp(raw)   & 0.2652    & 0.2530    & 5.39 & 0.3851        & 17.51  & 83.84    & 77.42   \\
        \hspace{0.3cm}- CodeExp(refined)   & 0.3175    & 0.3016    & 8.02 & \textbf{0.5536}        & 24.38  & 84.67    & 78.90   \\
        \hspace{0.3cm}- CodeExp(r+r) & \textbf{\underline{0.3415}}    & \textbf{\underline{0.3256}}    & \textbf{9.91} & \textbf{\underline{0.5695}}        & \textbf{\underline{26.87}}  & \textbf{\underline{84.98}}    & \textbf{\underline{79.52}}   \\
        \hline
        \end{tabular}
    }
    \caption{\label{table:result} Evaluate the generated explanation of fine-tuned models with various metrics. The \textbf{top three} scores for each metric are in bold text, and the \underline{top one} scores are underlined. }
    % \vspace{-0.3cm}
\end{table*}

\section{Results and discussion}
\label{sec:result}
% Results:
% a.	Showing results with auto-metrics. -> insights: (1) larger model, better performance (2) finetuning on (1,2) show significant improvements.
% b.	Human evals collected for 180 examples of 9 models. Kendal’s tau: how metrics agree with human evals. 

% Key subsections in discussion:
% c.	 Fine-tuning on (1,2) generates the best performance.
% d.	Fine-tuning on (2) generates sub-optimal performance but is much faster. Showing the time consumption and total flops.
% e.	Case studies, positive and negative cases; How the local and global views of code are handled; may show some insights on the attention scores.
% f.	Comparing with Codex.
% g.	For metrics, which metrics are concordant with human-eval. Propose the combination of auto-metrics.

% We applied the aforementioned auto-metrics on all the generated docstrings. Table~\ref{table:result} listed the scores of baseline models.

\subsection{Results of auto-metric evaluations}
We evaluated the generated docstrings of all aforementioned models (in Section~\ref{sec:model}). We also included the \textit{CodeT5} checkpoint for code summarization, i.e. \textit{CodeT5-multi-sum}\footnote{\url{https://huggingface.co/Salesforce/codet5-base-multi-sum}}. The evaluated results are shown in Table~\ref{table:result}. In latter sections, we use the notion "[model name]-(raw), -(refined), -(r+r)" representing the model fine-tuned using the strategy S1, S2, and S3, respectively.

We observed all fine-tuned models have significant improvements over off-the-shell versions across all metrics, for example, the BLEU scores increased from below 0.4 (\textit{GPT-Neo13}, \textit{CodeT5-multi-sum}) to around 10.0 (\textit{GPT-Neo13-(r+r)}, \textit{CodeT5-(r+r)}). Since \textit{GPT-Neo} and \textit{CodeT5-multi-sum} are reported as strong summarization baselines \cite{wang2021codet5}, this result again highlights the large difference between summaries and explanatory docstrings.

\paragraph{Two-stage fine-tuning achieves best performances.} For the three fine-tuning strategies (\textbf{S1-3} in Section~\ref{sec:exp-setup}), S3 yields the best performance for all baseline models. To recall, it trains on all collected examples (i.e. CodeExp(raw)) and the high-quality partition (i.e. CodeExp(refined)) consecutively. Comparing across models, fine-tuned CodeT5 models perform best for S2 and S3 fine-tuning. Especially, \textit{CodeT5-(r+r)} achieves the highest scores with respect to 6 out of 7 metrics.

\begin{table}[h]
\centering
    \resizebox{0.48\textwidth}{!}{
        \begin{tabular}{lccccc}
        \hline
        \textbf{Model}       & \textbf{A1} & \textbf{A2} & \textbf{A3} & \textbf{A4} & \textbf{Overall} \\
        \hline
        Reference        & \textbf{3.617}        & \textbf{2.994}        & \textbf{3.628}         & \textbf{3.822}       & \textbf{3.515}   \\
        \hline
        Codex-Py2Doc    & 3.394        & \textbf{2.950}         & 3.378         & 3.556       & 3.319   \\
        Codex-Py2NL    & 2.489        & 2.528        & 2.922         & 2.683       & 2.656   \\
        GPT-2-base-(r+r)       & 2.972        & 2.883        & 3.406         & 3.539       & 3.200   \\
        GPT-Neo27-(r+r) & 3.417        & 2.811        & 3.444         & 3.589       & 3.315   \\
        GPT-Neo13-(r+r)    & 3.283        & 2.900        & 3.439         & \textbf{3.606}       & 3.307   \\
        CodeT5-(raw)      & 2.594        & 2.217        & 2.756         & 2.889       & 2.614   \\
        CodeT5-(refined)      & \textbf{3.489}        & \textbf{3.061}        & \textbf{3.572}         & \textbf{3.661}       & \textbf{3.446}   \\
        CodeT5-(r+r)     & \textbf{3.478}        & 2.933        & \textbf{3.517}         & 3.578       & \textbf{3.376}   \\
        \hline
        \end{tabular}
    }
    \caption{Aspect-wise and overall score of Human evaluations. Four aspects (Section~\ref{sec:human-eval}): \textbf{A1}. General adequacy, \textbf{A2.} Coverage, \textbf{A3.} Coherence, \textbf{A4.} Fluency.}
    \label{tab:human-eval-scores}
\end{table}

% 1. lar
\subsection{Results of human evaluations}
\label{sec:result-human-eval}
We randomly select 180 examples from the test set and collect human evaluations using the protocol in Section~\ref{sec:human-eval}. For each evaluated docstring, the annotator gives four scores for the aspects \textbf{A1-4} and an overall average score is computed as well. The evaluated docstrings are generated using 6 fine-tuned models given the selected 180 python functions. The models and respective results are shown in Table \ref{tab:human-eval-scores}. For comparison, we also include the human-written reference docstrings in the original codebases and two strong baselines using the OpenAI Codex API, i.e. \href{https://beta.openai.com/examples/default-python-docstring}{Codex-PyDoc} and \href{https://beta.openai.com/examples/default-python-to-natural-language}{Codex-Py2NL}. These two APIs generate docstrings and NL explanations, respectively, and we follow the official settings (Appendix~\ref{sec:codex-api}).

Comparing the overall scores in Table~\ref{tab:human-eval-scores}, the relative superiority of models are largely consistent with the results of auto-metrics (Table~\ref{table:result}): (1) The fine-tuning strategies (S2, S3) using high-quality data partition outperforms S1. In fact, we found that fine-tuning using only CodeExp(raw) would often generate short one-line texts (like summarization) due to the majority of one-line docstring in the data, and this explains the annotators' low rating to CodeT5-(raw). (2) CodeT5-(refined) and CodeT5-(r+r) outperform other models.

\begin{figure}[h]
    \centering
    \includegraphics[width=0.48\textwidth]{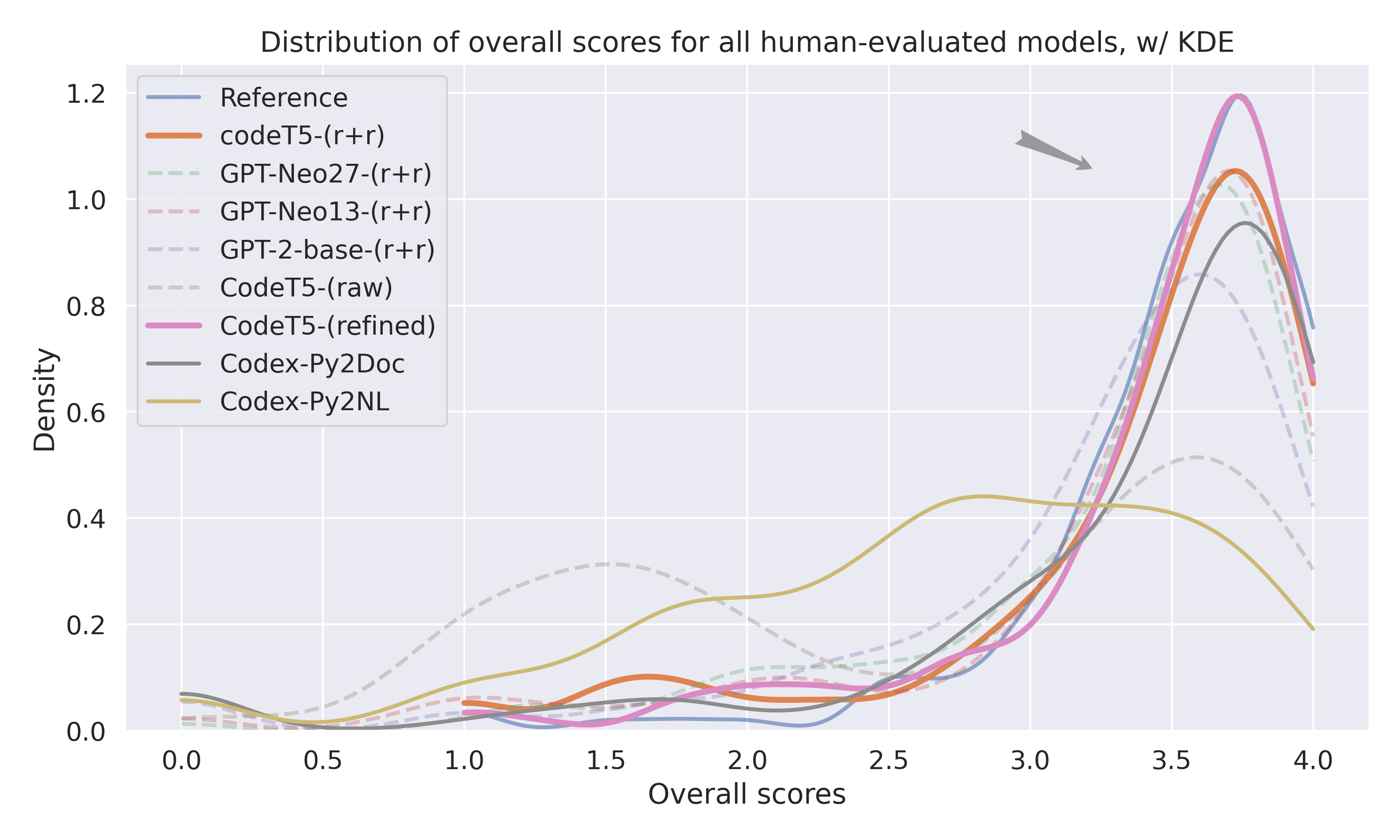}
    \caption{\textbf{Distribution of overall scores on 180 human-evaluated examples.} The best model has comparable performance with human-written reference docstrings.}
    \label{fig:kde}
\end{figure}

\paragraph{Achieving human-comparable performance.} We also observe that CodeT5-(refined) achieves comparable overall scores as the human-written references. Especially, it has a higher \textit{Coverage} score than the references, indicating more detailed explanations. We plot in Figure~\ref{fig:kde} the (kernel density estimated) distribution of overall scores across 180 examples. The CodeT5-(refined) has the highest density accumulated at the high score range (3.5-4.0), and the distribution is very close to the distribution of reference docstrings.

We also observe the best fine-tuned models outperform \textit{Codex-Py2Doc} and \textit{Codex-Py2NL} with respect to all evaluation aspects, although the parameter size of Codex (12B) is 60 times larger than the fine-tuned CodeT5.

\begin{table}[h]
\centering
    \resizebox{0.48\textwidth}{!}{
        \begin{tabular}{lccccc}
        \hline
        \textbf{Metric}        & \textbf{A1}    & \textbf{A2}    & \textbf{A3}    & \textbf{A4}    & \textbf{Overall} \\ 
        \hline
        ROUGE-1f      & 0.389          & 0.293          & 0.275          & 0.463          & 0.311            \\
        ROUGE-Lf      & 0.377          & 0.292          & 0.275          & 0.462          & 0.304            \\
        \textbf{BLEU}          & 0.416          & 0.347          & 0.327          & \textbf{0.554} & 0.355            \\
        \textbf{METEOR}        & \textbf{0.417} & 0.368          & \textbf{0.333} & 0.518          & \textbf{0.361}   \\
        BERTScore     & 0.278          & 0.218          & 0.251          & 0.471          & 0.263            \\
        CodeBERTScore & 0.334          & 0.266          & 0.280          & 0.494          & 0.297            \\
        \textbf{CER}     & 0.346          & \textbf{0.377} & 0.328          & 0.454          & 0.339            \\ 
        \hline
        \end{tabular}
    }
    \caption{Kendall's $\tau$ calculated between 7 automatic metrics and the human evaluated scores.}
    \label{tab:tau}
\end{table}

\paragraph{BLEU and METEOR are most consistent with human evaluations.} To examine the consistency between automatic metrics and human evaluation, we calculated the adapted Kendal's $\tau$ (Section~\ref{sec:human-eval}) and show the results in Table~\ref{tab:tau}. We found BLUE and METEOR mostly match the human evaluations for aspects A1,3,4 and the overall score. For the aspect of A2, Coverage, the newly introduced CER has the highest $\tau$. Interestingly, ROUGE scores show less alignment to human evaluations, although they have been widely used in code summarization. In summary, we recommend applying BLEU and METEOR to the task of code explanation generation.

\subsection{Data quality matters}
Reviewing both automatic and human evaluations, we find fine-tuning solely on the high-quality partition (CodeExp(refined)) significantly improves the performance compared to fine-tuning on CodeExp(raw). Taking as example the best performed model series, CodeT5: (1) For human evaluation, the CodeT5-(refined) achieves an overall score of 3.446 (ranking 1st) and improves 31.8\% over the score 2.614 of CodeT5-(raw) (Table~\ref{tab:human-eval-scores}). (2) For automatic metrics, BLEU and METEOR are the two most faithful metrics recommended in Section~\ref{sec:result-human-eval}. CodeT5-(refined) improves the BLEU score from 5.39 to 8.02 and METEOR score from 17.51 to 24.38, when compared to CodeT5-(raw). In fact, the preference of -(refined) over -(raw) can be observed for most automatic metrics and model types (Table~\ref{table:result}). This pattern is particularly interesting considering that the refined partition is only 1/15 of the total size. One reason is that the majority of CodeExp(raw) are of short length and do not satisfy the requirements of code explanation documents (Section~\ref{sec:data-select}). Therefore, it brings about observable noise during the optimization. This result demonstrates the importance of data quality for code explanation generation.

\subsection{Case Study}

Figure~\ref{fig:indent_example} shows an example of code and generated docstring. More examples are listed in Appendix~\ref{sec:example}. The CodeT5-(r+r) model successfully captures the main logic of "Indent text by a given number of characters". It also describes in detail the types and semantics of input parameters and returns. Interestingly, the model also captures the condition for the first "ValueError", although a more faithful description should be "ValueError: if the number of characters differs from the number of lines".

Notably, achieving human-written quality does not mean perfect. In this example, the human-written reference missed the description for the ValueErrors. We found the coverage aspect is challenging for both human-written and model-generated docstrings in general. This has also been reflected by the A2 scores in Table~\ref{tab:human-eval-scores}.
\begin{figure}[h]
    \centering
    \includegraphics[width=0.48\textwidth]{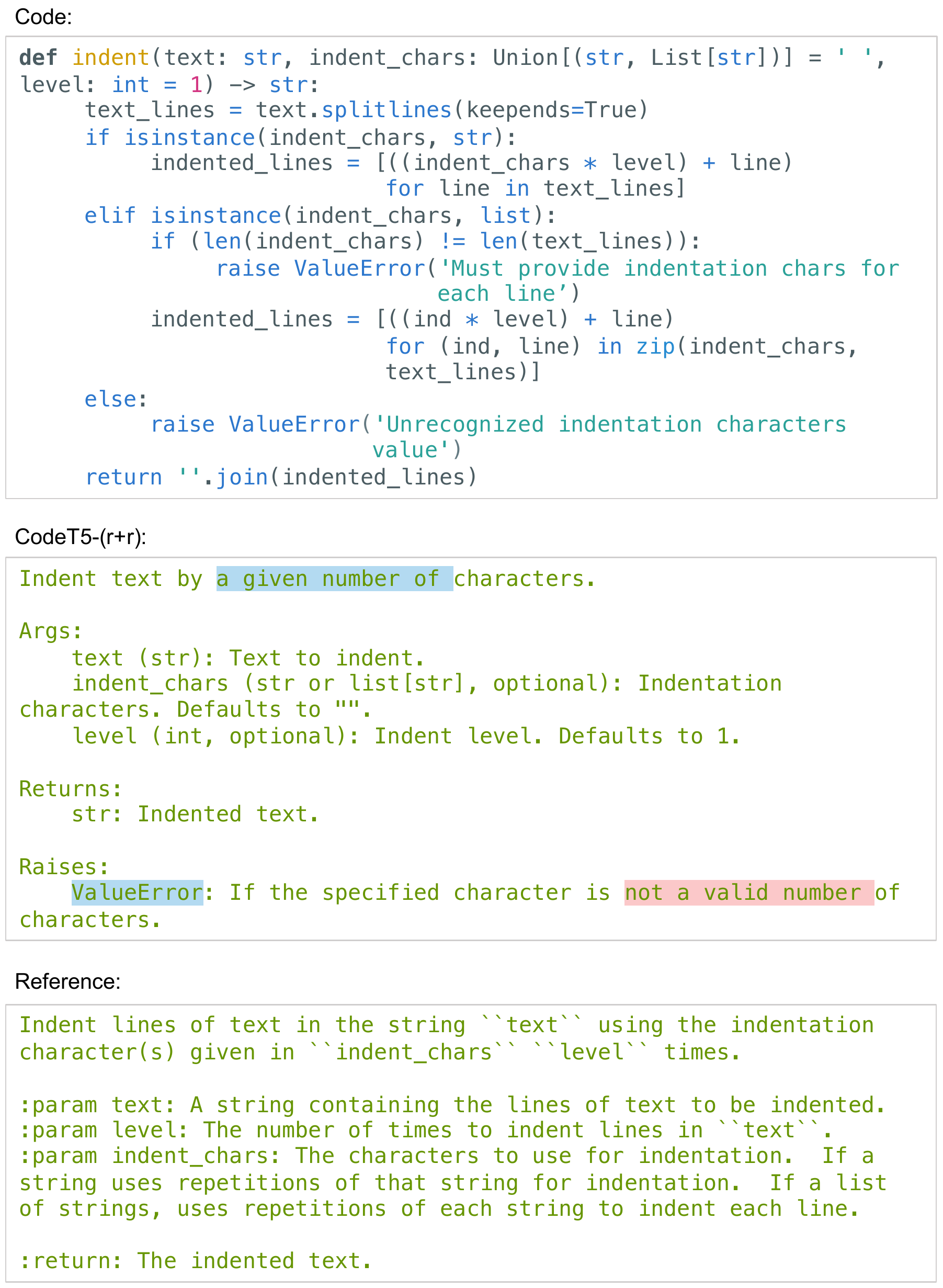}
    \caption{\textbf{Case study of generated docstring.} The CodeT5-(r+r) correctly captured some detailed code semantics (highlighted in \textcolor{custom_cyan}{cyan}). An ambiguous span is highlighted in \textcolor{custom_pink}{pink} as well.}
    \label{fig:indent_example}
\end{figure}

% \subsection{Discussion}

% Interestingly, it is not more data better performance style. In most of the experiments, fine-tuning with the high-quality data provides comparable performance, and even for the largest data we tested (GPT-Neo27) two-stage finetuning using all data and the high-quality partition provides marginal improvements comparing to the enlarged 10-time computation cost.

\section{Related works}

% \subsection{Code to Comment Translation}
% 1. method level automatic code documentation
% 2. decade ago works for full-length documentation
% 3. recent deep-learning based studies focus more on code summarization.
% 4. only a few do docstring generation, pymt5, codex
% 5. other level 
% 6. evaluation.

Several approaches have been proposed for method/function-level automatic code documentation. Early studies use template-based and information retrieval approaches to generate long-form documents \cite{wong2013autocomment, mcburney2014automatic, moreno2013jsummarizer}. Recent efforts have been mainly focused on the summarization. \citet{barone2017parallel} collected a large parallel Python code and docstring corpus and trained LSTM-based machine translation model. DeepCom \cite{hu2018deep} introduced structure information of AST to help generate summaries for Java methods. \citet{zhou2019augmenting} proposed ContectCC, which encodes the context information of external dependencies using the API calls in the source Java method. More recently, \citet{ahmad2020transformer} proposed a transformer approach for method level summarization. \citet{zhang2020retrieval} built a retrieval-based approach using similar code snippets in the generation. The aforementioned studies employed several popular PL-NL corpora \cite{barone2017parallel, hu2018deep, husain2019codesearchnet}. The average length of these corpora is below 20 tokens and they target the summarization task where the docstrings are often one-liners.
Apart from these efforts, few deep learning based approaches generate full-length of documentation. \citet{clement2020pymt5} pretrained T5 models to generate python docstring in numerous styles. The OpenAI Codex \cite{chen2021evaluating} trained GPT-3 for both code and document generation in six programming languages. 
% \citet{iyer2016summarizing} used attention-based networks to generate summaries for Stackoverflow snippets.

In these mentioned studies, machine translation (MT) metrics have been widely used for code comment assessments. \citet{gros2020code} questioned this adoption of reference-based MT metrics by examining and showing the semantic difference between code and NL. \citet{roy2021reassessing} rigorously examined the consistency between automatic metrics and human assessments using Kendall's $\tau$ as well and recommended the usage of BLEU, METEOR, and chrF.

\section{Conclusion}

The code explanation generation is an important task for code understanding. On one hand, we show existing summarization methods do not directly apply to this task. On the other hand, we built data collection pipelines, explored consistency between automatic and human evaluations, and provided a framework for fine-tuning existing pretrained models to generate explanatory docstrings of human-comparable quality. We highlight the importance of data quality by showing that fine-tuning on high-quality data exceeds the performance using raw data of 15 times larger scale. We expect the proposed infrastructures, including the annotated dataset, human-evaluation protocol, recommended metrics, and fine-tuning strategy, to boost future research for code explanation.

\section{Limitations}
The examined automatic metrics provide insufficient semantic verification for the generated docstrings. Also, the absolute $\tau$ values of automatic metrics are all below 0.5, which indicates limited consistency with respect to human evaluations. We look forward to potential factuality-based metrics better modeling the correctness and coverage of the explained semantics. Apart from evaluating stand-alone generations, a user study in the production/developer environment could more accurately reflect the effectiveness of AI-generated explanatory documents. As for the model performance, increasing the coverage over detailed code semantics remains a challenge for tested models. In fact, both generated and human-written docstrings have low coverage scores in the human assessments. Lastly, we tested various fine-tuning strategies in this work, while the large-scale pretraining for code explanation is as well worth exploring.

% Looking forward, 
% Unique challenge of global/local can do better with arch and method design.
% Improve coverage, style transfer.

% Also, the data-quality may be an open question, seeing the growing trends of using as more data as possible, we wonder the strategy could be further tested whether using all or emphasizing on the best-quality data.

% Style transfer in different scenarios.

% We only tested fine-tuning, it is worth testing explantion related pretraining objectives.

% scope of context, using class and project level information in the future would help

% \begin{table*}
% \centering
% \begin{tabular}{lll}
% \hline
% \textbf{Output} & \textbf{natbib command} & \textbf{Old ACL-style command}\\
% \hline
% & &  \\
% & &  \\
% & &  \\
% \hline
% \end{tabular}
% \caption{\label{citation-guide}
% Citation commands supported by the style file.
% The style is based on the natbib package and supports all natbib citation commands.
% It also supports commands defined in previous ACL style files for compatibility.
% }
% \end{table*}

\bibliography{anthology,custom}
\bibliographystyle{acl_natbib}

\newpage

\appendix
\renewcommand\thefigure{\thesection\arabic{figure}}

\section{Appendices}
\label{sec:appendix}

\subsection{Statistics of CodeExp annotation results}
\label{sec:ann-stats}

\begin{figure}[h]
    \centering
    \includegraphics[width=0.45\textwidth]{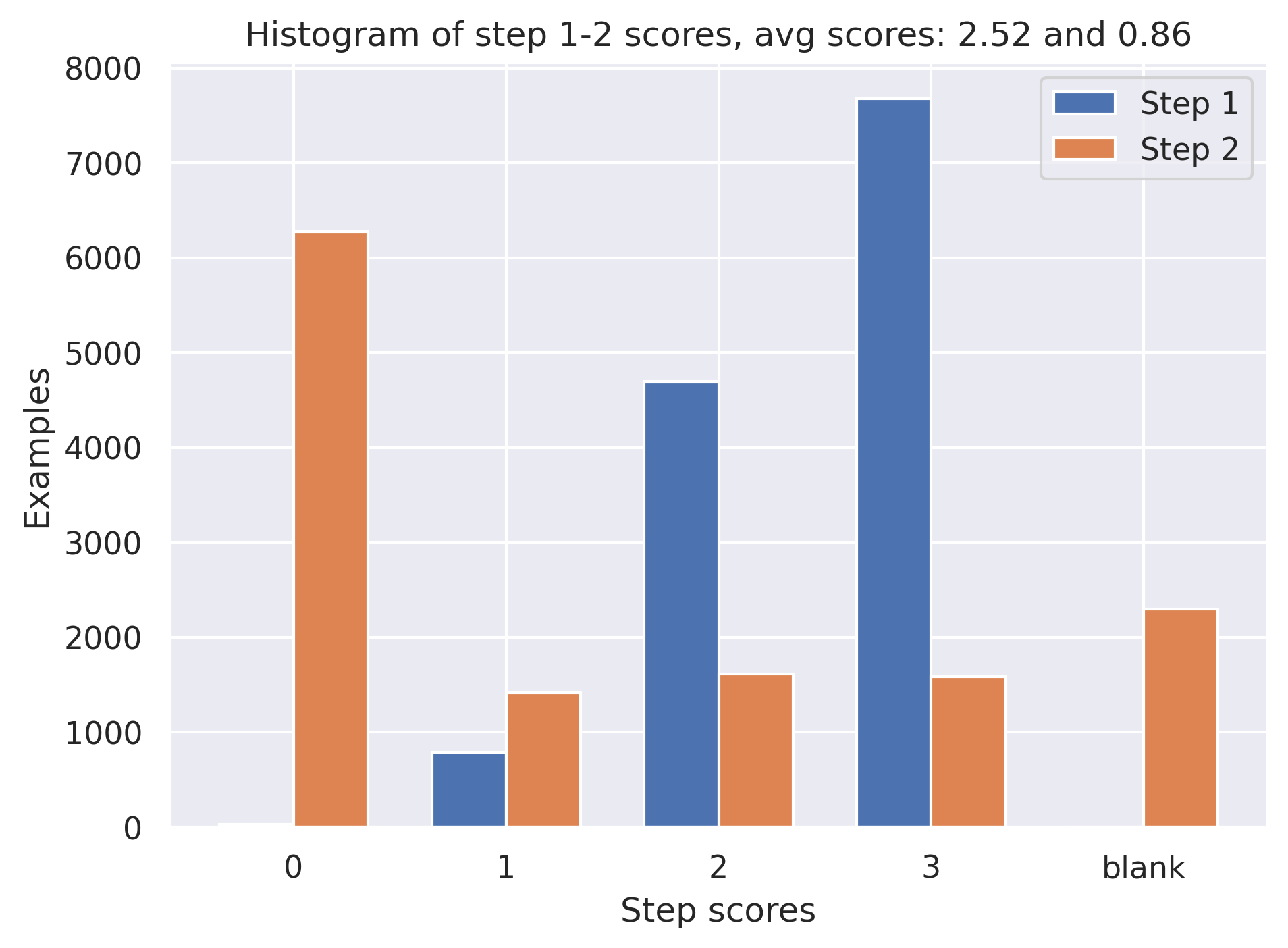}
    \caption{Histogram of Step 1. General adequacy and Step 2. Coverage. Blank score of step 2 indicates there is no branching if/else conditions in the code example.}
    \label{fig:step2_hist}
\end{figure}

\subsection{Test set configuration}
\label{sec:test-set}
The test set mainly consists of examples in the CodeExp(annotated) partition. We select the group of data of high quality, i.e. with scores of all steps $\geq 1$ (including blank scores), and remove duplicates that also appeared in the CodeExp(raw) and CodeExp(refined). This process generates 1744 examples. We also applied the same procedure of quality filtering and deduplication on a small held-out set of GitHub code-doc pairs. Altogether the test set contains 2677 examples.

In detail, the deduplication works by computing the Levenshtein distance between a candidate string of code/document and each code/document in the 2.3 million CodeExp(raw). To accelerate, We compare the first 300 characters of the candidate and target strings. If any computed distance is less than 5\% of the total string length, the candidate code-doc example will be considered a duplicate and excluded from the test set.

\begin{table*}[h]
    \centering
    \resizebox{0.97\textwidth}{!}{
        \begin{tabular}{lccccc}
        \hline
        \textbf{Model} & \textbf{max input tokens} & \textbf{optimizer}     & \textbf{learning rate} & \textbf{batch size} & \textbf{\#epochs} \\
        \hline
        GPT-2-base-(raw) & 1024                      & \multirow{8}{*}{AdamW} & \multirow{8}{*}{\makecell[c]{start at 5e-5, \\linear decay to 0}}     & 16                  & 3                 \\
        GPT-2-base-(refined) & 1024                      &                        &                        & 16                  & 3                 \\
        GPT-Neo13-(raw)  & 2048                      &                        &                        & 16                  & 3                 \\
        GPT-Neo13-(refined)  & 2048                      &                        &                        & 16                  & 3                 \\
        GPT-Neo27-(raw)  & 2048                      &                        &                        & 16                  & 3                 \\
        GPT-Neo27-(refined)  & 2048                      &                        &                        & 16                  & 3                 \\
        CodeT5-(raw)     & 1024                      &                        &                        & 32                  & 10                \\
        CodeT5-(refined)     & 1024                      &                        &                        & 32                  & 10               \\
        \hline
        \end{tabular}
    }
    \caption{Fine-tuning hyperparameters. Note: (1)All max tokens are set to the upper limits of the model\_max\_token of the pretrained model from \url{Huggingface.co} (2)The two-stage fine-tuning for each model adopts the settings of -(raw) and -(refined) at each stage, respectively.}
    \label{tab:hyper-param}
\end{table*}

\subsection{Inference settings}
\label{sec:inference}
At inference time, the generated tokens are sampled with temperature set to 0.1 for all models. The max generated token is set according to each model's capacity. Specifically, 512 tokens for CodeT5 and GPT-Neo models, 256 tokens for GPT-2-base.

\subsection{Human evaluation settings}
\label{sec:human-eval-setting}
The human evaluation consists of four aspects. For each aspect, a question related to the docstring quality is asked, and the annotator is expected to give an integer score within 0-4, where 0 stands for "not satisfying the question at all" and 4 stands for "perfectly satisfying the question". The four aspects and questions are as follows:
\begin{enumerate}
    \item General adequacy: Is it possible to gain a basic understanding of what the code does after reading the docstring?
    \item Coverage: How much does the docstring cover important semantic details, including descriptions for input parameters, returns, exceptions and if/else, try/catch blocks?
    \item Coherence: Is the information provided in the docstring correct and related to the code?
    \item Fluency: Is the docstring grammatically correct and easy to read?
\end{enumerate}
Notably, the first three aspects reflect those aspects of data annotations in Section~\ref{sec:task}.

Given one source code, each annotator must evaluate the generated docstrings of all models (including the reference docstring) to remove inter-model bias. The source of the docstring is hidden to the annotators. In total, ten annotators provided 6480 scores, 180 examples $\times$ 9 models $\times$ 4 aspects. All annotators have python developing experience of over 2 years.

\subsection{Codex API settings}
\label{sec:codex-api}

The Codex-Py2Doc stands for the Codex API example of "\href{https://beta.openai.com/examples/default-python-docstring}{Write a Python docstring}". The official prompt includes the \textit{\# Python 3.7} header, the source code, and appends at the end the prompting sentence \textit{\# An elaborate, high quality docstring for the above function: """}. The model stops generating when the stop token \textit{\#} or \textit{"""} is generated. 
Similarly, the Codex-Py2NL denotes the Codex API of "\href{https://beta.openai.com/examples/default-python-to-natural-language}{Python to natural language}". The prompt includes the \textit{\# Python 3} header, the source code, and the prompting line \textit{\# Explanation of what the code does \textbackslash n \textbackslash n \#}.

We follow all official settings for these APIs, including setting the temperature to 0, top p to 1.0, frequency penalty to 0.0, and presence penalty to 0.0. We use the most capable engine available, "code-davinci-002" and increase the max generated tokens to 256. The default stop token \textit{\#} for Codex-Py2NL is removed in our settings because otherwise the API would only generate one line of text.

\subsection{Examples of generated docstrings}
\begin{figure*}[h!]
    \centering
    \includegraphics[width=\textwidth]{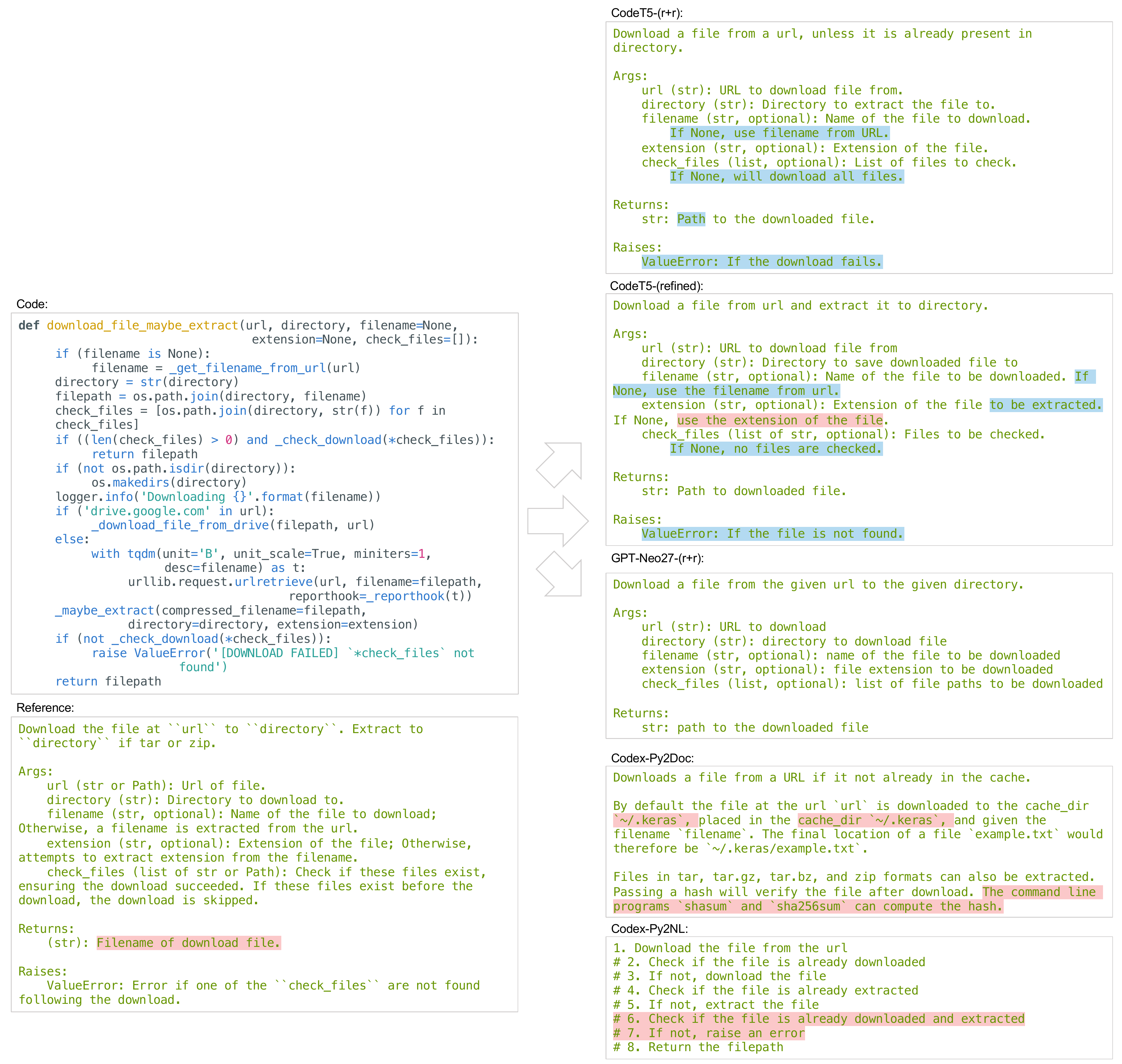}
    \caption{An example function, human-written reference docstring, and selected generated docstrings. The best fine-tuned models captures both global and local function behaviors. Highlighted text: \textcolor{custom_pink}{pink} - notable errors in generation or reference. \textcolor{custom_cyan}{cyan} - key information captured in generation.}
    \label{fig:download_file_example}
\end{figure*}

See Figure~\ref{fig:download_file_example}.
\label{sec:example}

% Interesting findings:

% 1. Codes with proper type defs and default values usually can help generate better docstrings. \textbf{Show statistics} In general, with better code quality, model generates better docstrings. 

% 2. Show the correlation of generation scores with respect to the original human-eval scores for the code-doc pairs. Particularly the general logic scores and the coverage scores.

\end{document}